\documentclass[orivec]{llncs}
\usepackage[T1]{fontenc}
\usepackage{graphicx}
\usepackage{amsmath}
\usepackage{amssymb}
\usepackage{verbatim}
\usepackage{booktabs}
\usepackage{multirow}
\makeatletter
\renewcommand\section{\@startsection{section}{1}{\z@}%
                       {-8\p@ \@plus -2\p@ \@minus -2\p@}%
                       {4\p@ \@plus 1\p@ \@minus 1\p@}%
                       {\normalfont\large\bfseries\boldmath
                        \rightskip=\z@ \@plus 8em\pretolerance=10000 }}

\renewcommand\subsection{\@startsection{subsection}{2}{\z@}%
                       {-6\p@ \@plus -2\p@ \@minus -2\p@}%
                       {3\p@ \@plus 1\p@ \@minus 1\p@}%
                       {\normalfont\normalsize\bfseries\boldmath
                        \rightskip=\z@ \@plus 8em\pretolerance=10000 }}

\renewcommand\subsubsection{\@startsection{subsubsection}{3}{\z@}%
                       {-5\p@ \@plus -1\p@ \@minus -1\p@}%
                       {-0.35em \@plus -0.12em \@minus -0.05em}%
                       {\normalfont\normalsize\bfseries\boldmath}}
\makeatother
\usepackage{enumitem}

\setlist[itemize]{
  topsep=2pt,
  itemsep=1pt,
  parsep=0pt,
  partopsep=0pt,
  leftmargin=1.2em
}

\setlist[enumerate]{
  topsep=2pt,
  itemsep=1pt,
  parsep=0pt,
  partopsep=0pt,
  leftmargin=1.4em
}
\usepackage[most]{tcolorbox}
\usepackage[table]{xcolor}
\AtBeginDocument{
  \setlength{\abovedisplayskip}{3pt}
  \setlength{\belowdisplayskip}{3pt}
  \setlength{\abovedisplayshortskip}{1pt}
  \setlength{\belowdisplayshortskip}{1pt}
}

\definecolor{BestGray}{HTML}{E6E6E6}
\definecolor{BoxGray}{HTML}{4A4A4A}
\definecolor{BoxLightGray}{HTML}{F7F7F7}

\newtcolorbox{graybox}{
  enhanced,
  breakable,
  colback=BoxLightGray,
  colframe=BoxLightGray,
  borderline west={2pt}{0pt}{BoxGray},
  boxrule=0pt,
  arc=0pt,
  left=7pt,
  right=5pt,
  top=1pt,
  bottom=1pt,
  before skip=2pt,
  after skip=2pt,
}
\makeatletter
\def\@spthm#1#2#3#4{%
  \topsep 3\p@ \@plus1\p@ \@minus1\p@ 
  \refstepcounter{#1}%
  \@ifnextchar[{\@spythm{#1}{#2}{#3}{#4}}{\@spxthm{#1}{#2}{#3}{#4}}}
\makeatother

\definecolor{ParaGray}{HTML}{F1F1F1}

\newcommand{\mypara}[1]{%
  \noindent\colorbox{ParaGray}{\textit{\textbf{#1}}}
}
\begin{document}
%
\title{Quantum-Structured World Models (QSWMs) for Predictive Latent Dynamics}
\titlerunning{Quantum-Structured World Models}
%
\author{
Hailong Jiang\inst{1}\thanks{Corresponding author.} \and
Emran Hossain\inst{1} \and
Feng Yu\inst{1} \and
Jianfeng Zhu\inst{2} \and
Guilin Zhang\inst{3} \and
Wulan Guo\inst{3}
}

\institute{
Youngstown State University, Youngstown, OH, USA\\
\email{\{hjiang, fyu\}@ysu.edu}, \email{ehossain@student.ysu.edu}
\and
Kent State University\\
\email{jzhu10@kent.edu}
\and
George Washington University\\
\email{\{guilin.zhang, wulan.guo\}@gwu.edu}
}
%
%
%
\maketitle              
 \begin{abstract}
World models learn latent states that summarize interaction histories, evolve over time, and support prediction, simulation, or planning. Most existing world models represent these states using classical vectors, probability distributions, recurrent hidden states, or transformer activations. In this paper, we introduce \emph{Quantum-Structured World Models} (QSWMs), a quantum-inspired framework for predictive world modeling with structured latent states, latent transition operators, and measurement-inspired decoding maps. We study whether mathematical structures inspired by quantum theory, such as complex-valued representations and density-matrix-like latents, provide useful inductive biases for world modeling. We establish three foundational properties: classical inclusion, predictive sufficiency, and structured compactness. We then instantiate complex-valued and density-matrix-like QSWM variants and evaluate them on elementary cellular automata against strong classical baselines. Results show promising local predictive potential for complex-valued QSWMs, while also revealing limitations in long-horizon rollout, density-matrix variants, and latent interpretability.


\end{abstract}

\section{Introduction}
\label{sec:introduction}

\mypara{Background.} World models have become a central abstraction for prediction and planning~\cite{ha2018worldmodels,hafner2019planet}.
Rather than directly mapping inputs to outputs, a world model learns a latent state that summarizes past observations and actions, evolves through learned dynamics, and supports future prediction, simulation, or planning.
This latent-state view has shaped model-based reinforcement learning, embodied AI, robotics, and latent dynamics learning~\cite{hafner2020dreamer,hafner2021dreamerv2,hafner2023dreamerv3,schrittwieser2020muzero,kaiser2020simple}.

The key object in world models is the latent world state \cite{kim2026latent}.
Most existing models represent this state as a real-valued vector, probability distribution, recurrent hidden state, or transformer activation~\cite{sutton2018reinforcement,kaelbling1998planning,vaswani2017attention,cho2014learning,hochreiter1997long}.
These representations are powerful, but they mainly encode latent information as features or probabilities.
They do not explicitly model quantum-inspired structures such as complex amplitudes, density-like operators, or coherence-like relations among latent alternatives \cite{encyclopedia5020048}.
In dynamical systems, future states may depend not only on which latent alternatives are possible, but also on how they interact under the learned dynamics \cite{hazan2025research}.

This motivates quantum-structured latent states as classical, quantum-inspired representations for organizing latent possibilities, correlations, and structured interactions in world models \cite{maio2026representational}.
Quantum theory provides a useful mathematical language for amplitudes, uncertainty, coherence-like relations, and measurement-based readout~\cite{schuld2015introqml,biamonte2017quantum}.
These ideas are relevant because world-model prediction often depends on multiple latent alternatives and their interactions under learned dynamics \cite{huang2025ladi,yao2025navmorph,gao2025adaworld}.
Inspired by QML and QRL studies on quantum states, parameterized quantum circuits, quantum search, and quantum-inspired learning~\cite{schuld2015introqml,biamonte2017quantum,zeng2023quantumworldmodel,zeng2024quantumtd,zhang2026readout}, QSWMs introduce quantum-structured latent representations into predictive world modeling.

This paper introduces \emph{Quantum-Structured World Models} (QSWMs), a quantum-inspired framework for studying world models with quantum-structured latent states.
A QSWM maps an interaction history $h_t$ to a quantum-structured latent state, evolves this state through a latent transition operator, and extracts predictions through a measurement-inspired decoding map:
\[
\rho_t=E_\theta(h_t), \qquad
\rho_{t+1}=\mathcal{T}_{\theta,a_t}(\rho_t), \qquad
\hat{o}_{t+1}=\mathcal{M}_{\phi}(\rho_{t+1}).
\]
Here, quantum-structured latent states are implemented in classical neural architectures using structures inspired by quantum theory, such as complex-valued representations and density-matrix-like latent states \cite{zhang2022complex,shi2021two}.
This makes QSWMs a representation-level framework for studying how quantum-inspired latent structures affect predictive world modeling.
The central question is:
\begin{graybox}
\textit{whether quantum-structured latent representations provide useful inductive biases for predictive world modeling and offer a representational basis for studying potential emergent world-modeling capabilities.}
\end{graybox}

We address this question through formal analysis and controlled empirical validation.
We first define QSWMs as formal objects for latent world modeling and establish three foundational properties: classical inclusion, predictive sufficiency, and structured compactness.
These properties show that classical probabilistic world models can be recovered as restricted QSWMs, that QSWM latent states can be evaluated through future-relevant predictive sufficiency, and that quantum-structured representations can compactly express certain correlated dynamics.

We then instantiate two quantum-inspired variants, ComplexQSWM and DensityQSWM, and evaluate them on elementary cellular automata against capacity-matched, doubled-latent, and normalized classical baselines.
Our results demonstrate the promise of QSWMs as predictive latent world models.
ComplexQSWM achieves the strongest local predictive performance among all evaluated models and consistently outperforms capacity-matched, doubled-latent, and normalized classical baselines.
Scaling experiments further show that ComplexQSWM maintains favorable trends as latent capacity increases, and out-of-distribution density tests indicate that its benefit extends beyond a single fixed data setting.
These findings suggest that quantum-structured latent representations can provide useful inductive bias for local predictive dynamics and offer a promising representational basis for studying emergent world-modeling capabilities.
This paper makes the following contributions:
\begin{itemize}
\item We introduce \emph{Quantum-Structured World Models} (QSWMs), a quantum-inspired framework for predictive world modeling with quantum-structured latent states, latent transition operators, and measurement-inspired decoding.
\item We establish three foundational properties of QSWMs: classical inclusion, predictive sufficiency, and structured compactness.
\item We instantiate ComplexQSWM and DensityQSWM and provide controlled empirical evidence that quantum-structured latent representations can offer useful local predictive inductive bias beyond strong classical controls.
\end{itemize}

\section{Background and Related Work}
\label{sec:background}

QSWM builds on three related areas: QML, world models, and emergent ability.\\
\mypara{QML and QRL.}
QML studies how quantum states, circuits, measurements, and quantum-inspired representations can support learning~\cite{nielsen2010quantum,schuld2015introqml,biamonte2017quantum,schuld2018supervised}.
QRL extends these ideas to decision-making and control.
Recent work has also introduced quantum world models in quantum reinforcement learning~\cite{zeng2023quantumworldmodel,zeng2024quantumtd}.
QSWM differs in emphasis.
Its primary object is not a classifier, regressor, value function, policy, or search procedure.
Instead, QSWM focuses on the internal world model itself: how histories are encoded into quantum-structured latent states, how those states evolve, and how predictions or imagined futures are extracted.\\
\mypara{World models} learn internal latent dynamics for prediction, simulation, and planning.
They have been widely used in model-based reinforcement learning and latent imagination~\cite{ha2018worldmodels,hafner2019planet,hafner2020dreamer,hafner2021dreamerv2,hafner2023dreamerv3,schrittwieser2020muzero,kaiser2020simple,sutton2018reinforcement}.
A typical world model encodes observations and actions into a latent state, evolves this state with learned dynamics, and decodes future observations or rewards.
Most existing world models use classical latent representations, such as vectors, probability distributions, recurrent states, or transformer activations~\cite{vaswani2017attention,sutskever2008recurrent}.
QSWM extends this abstraction by allowing the latent world state and its transition dynamics to be quantum-native or quantum-inspired.\\
\mypara{Emergent abilities} refer to capabilities that are weak or absent in simpler models but become visible when model scale, training, or representational capacity increases~\cite{berti2025emergent}.
In recent AI systems, this term is often used to describe abilities such as reasoning, in-context learning, instruction following, and multi-step problem solving that appear or become much stronger beyond certain model scales. 
For world models, emergence raises a central question: whether a learned latent state can support not only local prediction, but also higher-level capabilities such as internal simulation, abstraction, and planning.


\section{Quantum-Structured World Models}
\label{sec:definition}

\begin{figure}[htbp]
    \centering
    \includegraphics[width=0.9\linewidth]{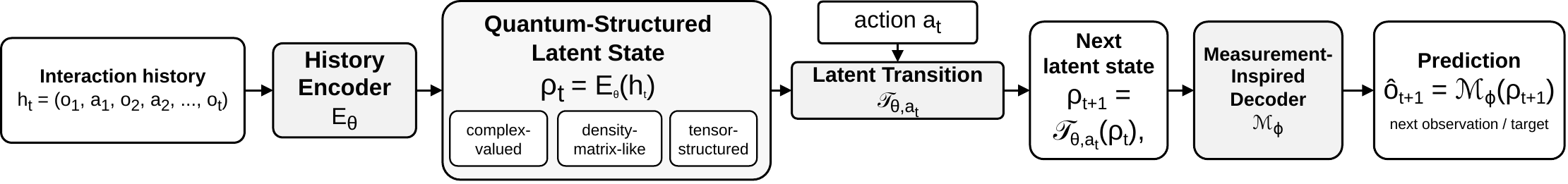}
    \caption{\small The overview of QSWM.}
    \label{fig:overview}
\end{figure}

Consider an agent interacting with an environment over discrete time. 
At time $t$, the agent receives an observation $o_t\in\mathcal{O}$, may take an action $a_t\in\mathcal{A}$, and has access only to the interaction history
\[
h_t=(o_1,a_1,o_2,a_2,\ldots,o_{t-1},a_{t-1},o_t),
\]
rather than the true environment state $s_t$.

A classical latent world model encodes this history into a latent vector, evolves it through learned dynamics, and decodes future predictions~\cite{ha2018worldmodels,hafner2019planet,hafner2020dreamer}:
\[
z_t=E_\theta(h_t),\qquad
z_{t+1}=F_{\theta,a_t}(z_t),\qquad
\hat{o}_{t+1}=G_\phi(z_{t+1}).
\]

As shown in Figure~\ref{fig:overview}, a QSWM replaces the classical latent vector $z_t$ with a quantum-structured latent state $\rho_t$.
Here, ``quantum-structured'' refers to classical neural representations inspired by mathematical structures from quantum theory, such as complex-valued latent states, density-matrix-like latent states, and tensor-structured representations.
The goal is to introduce richer latent structure into world modeling, not to require quantum hardware execution.


Formally, $\rho_t$ can be viewed as an element of a density-operator-like space $\mathcal{D}(\mathcal{H})$ over a Hilbert space $\mathcal{H}$.
Our experiments instantiate this with two differentiable representations, including complex-valued and density-matrix-like latents.\\
\mypara{Definition 1: Quantum-Structured World Model.} A quantum-structured world model is a quantum-inspired world model that represents its latent world state using mathematical structures inspired by quantum theory.
A QSWM is a tuple
\[
\mathcal{W}_{\mathrm{QS}}=(E_\theta,\mathcal{T}_\theta,\mathcal{M}_\phi,\mathcal{L}),
\]
where $E_\theta$ encodes an interaction history into a quantum-structured latent state, $\mathcal{T}_{\theta,a_t}$ evolves this state under action $a_t$, $\mathcal{M}_\phi$ extracts classical predictions through a measurement-inspired decoding map, and $\mathcal{L}$ is a predictive or control-based training objective:
\[
\rho_t=E_\theta(h_t),\qquad
\rho_{t+1}=\mathcal{T}_{\theta,a_t}(\rho_t),\qquad
\hat{o}_{t+1}=\mathcal{M}_\phi(\rho_{t+1}).
\]
In this paper, ``quantum-structured'' refers to classical neural representations inspired by quantum theory, including complex-valued latent states and density-matrix-like latent states.
Thus, QSWM is a representation-level framework for studying how quantum-structured latent representations affect predictive world modeling.\\
\mypara{Definition 2: Predictive Sufficiency.}
A latent state $\rho_t=E_\theta(h_t)$ is predictively sufficient for a future target $y_t$ if
\[
p(y_t\mid h_t)=p(y_t\mid \rho_t).
\]
This condition states that $\rho_t$ preserves the information in the history that is needed for predicting $y_t$.
It provides a basic criterion for local world modeling and a foundation for studying stronger world-modeling capabilities such as long-horizon prediction, abstraction, simulation, and planning.

\section{Foundational Properties}
\label{sec:foundations}

This section summarizes three foundational properties of QSWMs.
These properties establish QSWMs as a grounded extension of classical latent world models, provide a predictive criterion for evaluating learned latent states, and show how quantum-structured representations can compactly express certain correlated dynamics.
Detailed proofs are provided in Appendix~\ref{app:proofs}.

\mypara{Theorem 1: Classical Inclusion.}
Every finite-dimensional classical probabilistic latent world model can be represented as a restricted QSWM whose latent states are diagonal density operators and whose transition and measurement maps are classical stochastic maps.
\noindent
\emph{Intuition.}
A classical belief vector $p_t=(p_t(1),\ldots,p_t(n))$ can be embedded as
\[
\rho_t=\sum_{i=1}^{n}p_t(i)|i\rangle\langle i|.
\]
Classical stochastic transitions preserve diagonality, and classical observation models correspond to measurements in the same basis. 
Thus, classical world models are recovered as diagonal QSWMs.\\
\mypara{Theorem 2: Predictive Sufficiency.}
If the encoder $E_\theta$ maps each interaction history $h_t$ to a latent state $\rho_t=E_\theta(h_t)$ such that
\[
p(y_t\mid h_t)=p(y_t\mid \rho_t),
\]
then $\rho_t$ is sufficient for predicting the future target $y_t$ from the history.

\noindent
\emph{Intuition.}
The latent state $\rho_t$ is useful when it preserves the information in $h_t$ that is needed for future prediction. 
This property does not claim emergent ability by itself. 
Rather, it gives a basic condition for local predictive world modeling: the model should compress the history into a latent state without losing future-relevant information.

\mypara{Theorem 3: Structured Compactness.}
Consider environments whose predictive state distributions over $n$ latent factors admit tensor-network representations with bond dimension at most $\chi$. 
A QSWM using tensor-network latent states can represent these predictive states with $O(n\chi^2)$ parameters, while an unstructured classical state table requires $O(2^n)$ parameters~\cite{orus2014tensor}.

\noindent
\emph{Intuition.}
An unstructured distribution over $n$ binary latent factors requires one value for each configuration in $\{0,1\}^n$. 
If the predictive distribution admits a bounded-bond-dimension tensor-network factorization, it can be represented by local tensors with $O(n\chi^2)$ parameters. 
This does not imply superiority over all classical neural models; it identifies a testable compactness condition for structured dynamics.


\section{QSWM Instantiations}
\label{sec:instantiation}
The QSWM framework supports multiple levels of implementation, including quantum-inspired, hybrid quantum-classical, and quantum-native designs.
This paper focuses on quantum-inspired instantiations implemented within classical differentiable architectures.
This setting provides a controlled testbed for studying how quantum-structured latent representations affect predictive world modeling, while keeping the training objective, data, and evaluation protocol comparable to classical world-model baselines.

We instantiate two lightweight QSWM variants: a complex-valued latent model (ComplexQSWM) and a density-matrix-like latent model (DensityQSWM).\\
\mypara{Complex-Valued QSWM (ComplexQSWM)}
represents the latent world state using explicit real and imaginary components. 
Given a history $h_t$, an encoder produces two real vectors:
\[
u_t, v_t = E_\theta(h_t),
\]
where $u_t,v_t \in \mathbb{R}^d$. 
They define a complex-valued latent state
\[
\psi_t = u_t + i v_t.
\]
To encourage a quantum-state-like representation, the latent state is normalized:
\[
\psi_t \leftarrow \frac{\psi_t}{\|\psi_t\|_2+\epsilon},
\]
where $\epsilon$ is a small numerical constant.

The transition operator is implemented as a learned complex-valued linear transformation represented by real-valued parameters. 
Writing $A=A_r+iA_i$, the transition
\[
\psi_{t+1}=A\psi_t
\]
is computed through real and imaginary components:
\[
u_{t+1}=A_r u_t - A_i v_t,
\qquad
v_{t+1}=A_r v_t + A_i u_t.
\]
The next-state prediction is decoded from the concatenated real and imaginary components:
\[
\hat{o}_{t+1}
=
D_\phi([u_{t+1},v_{t+1}]).
\]

This model is quantum-inspired rather than quantum-native. 
It does not require quantum hardware, but it introduces amplitude-like latent structure and norm-constrained latent evolution. 
The purpose is to test whether such structure can improve predictive world modeling under controlled conditions.\\
\mypara{Density-Matrix-Like QSWM (DensityQSWM)}
uses a density-matrix-like latent representation. 
Given a history $h_t$, an encoder produces a real vector
\[
v_t = E_\theta(h_t), \qquad v_t \in \mathbb{R}^d.
\]
The latent state is constructed as a positive semidefinite matrix:
\[
\rho_t
=
\frac{v_t v_t^\top}{\mathrm{Tr}(v_t v_t^\top)+\epsilon}.
\]
This construction ensures that $\rho_t$ is normalized and positive semidefinite up to numerical precision.

A learned transition matrix $A_\theta$ evolves the latent state:
\[
\tilde{\rho}_{t+1}
=
A_\theta \rho_t A_\theta^\top,
\]
followed by trace normalization:
\[
\rho_{t+1}
=
\frac{\tilde{\rho}_{t+1}}
{\mathrm{Tr}(\tilde{\rho}_{t+1})+\epsilon}.
\]
The flattened matrix $\rho_{t+1}$ is then decoded into next-state logits:
\[
\hat{o}_{t+1}
=
D_\phi(\mathrm{vec}(\rho_{t+1})).
\]
This model approximates the structure of density-state evolution while remaining fully classical and differentiable. 
It provides a simple test of whether positive semidefinite latent-state structure benefits predictive dynamics modeling.


Both models receive the same interaction history $h_t$, predict the next observation or state, and support recursive rollout evaluation.
They share the same encoder-transition-decoder pipeline, but differ in how the latent state $\rho_t$ is represented, normalized, and transformed.
More advanced QSWM variants may use tensor-network states, hybrid quantum-classical circuits, local measurement objectives, or quantum-native transition channels. 
We leave these extensions to future work 
in Appendix~\ref{app:architectures}.

\section{Experimental Validation}
\label{sec:experiments}

We evaluate QSWMs as predictive latent world models on controlled dynamical-system benchmarks.
The experiments are designed to answer four questions: whether QSWMs can be trained effectively, whether quantum-structured latent states improve local prediction beyond strong classical controls, whether the benefit persists under scaling and distribution shift, and how rollout and probing characterize the learned latent dynamics.

\mypara{Setup.}
We use one-dimensional elementary cellular automata as controlled latent-dynamics benchmarks.
Each state is a binary vector of length $L=32$.
Given a history of $H=4$ states, the model predicts the next cellular-automaton state.
Unless otherwise stated, we evaluate Rule 30, Rule 90, and Rule 110 with $5000/1000/1000$ train/validation/test trajectories per rule.
All models are trained for $20$ epochs using Adam with learning rate $10^{-3}$ and binary cross entropy.
Default initial states are sampled from a Bernoulli distribution with density $p=0.5$.

\mypara{Models and metrics.}
We compare \textbf{ComplexQSWM} and \textbf{DensityQSWM} with several classical controls: \textbf{ClassicalWorldModel}, parameter-matched classical baselines, \textbf{RealDoubledMatched}, and \textbf{RealNormalizedMatched}.
These controls test whether the observed gains are explained by parameter count, doubled real latent dimensions, or latent normalization.
We report one-step BCE and bit accuracy for next-state prediction. 
We also evaluate recursive rollout accuracy, which tests whether a learned world model can simulate future states by feeding its own predictions back as inputs~\cite{ha2018worldmodels,hafner2020dreamer}. 
To test robustness under distribution shift, we evaluate out-of-distribution (OOD) initial-density generalization~\cite{liu2021oodsurvey}. 
We further report scaling over latent dimensions $d\in\{16,32,64,128\}$ and latent probing with frozen world-model representations.

\begin{table}[t]
\centering
\caption{Main predictive results across Rule 30, Rule 90, and Rule 110.} 
\label{tab:main_results}
\scriptsize
\setlength{\tabcolsep}{3.5pt}
\begin{tabular}{lrrrrrrr}
\toprule
Model & Params & BCE $\downarrow$ & Acc@1 $\uparrow$ & Acc@3 $\uparrow$ & Acc@5 $\uparrow$ & Acc@10 $\uparrow$ & Probe $\uparrow$ \\
\midrule
ClassicalWM     & 37.2K & 0.5935 & 0.6654 & 0.5561 & 0.5281 & 0.5143 & 0.7812 \\
Matched-Complex & 39.3K & 0.5738 & 0.6806 & 0.5633 & 0.5285 & 0.5119 & 0.7866 \\
Matched-Density & 55.5K & 0.5681 & 0.6807 & 0.5654 & \cellcolor{BestGray}\textbf{0.5306} & \cellcolor{BestGray}\textbf{0.5151} & \cellcolor{BestGray}\textbf{0.7988} \\
ComplexQSWM     & 39.3K & \cellcolor{BestGray}\textbf{0.4890} & \cellcolor{BestGray}\textbf{0.7364} & \cellcolor{BestGray}\textbf{0.5798} & 0.5275 & 0.5037 & 0.6824 \\
DensityQSWM     & 55.9K & 0.5877 & 0.6673 & 0.5388 & 0.5135 & 0.5065 & 0.6676 \\
\bottomrule
\end{tabular}
\end{table}

\mypara{Finding 1: ComplexQSWM improves local predictive dynamics.}
Table~\ref{tab:main_results} shows that ComplexQSWM achieves the strongest one-step predictive performance across the evaluated rules.
It obtains the lowest BCE, $0.4890$, and the highest Acc@1, $0.7364$, among all models.
This result shows that complex-valued quantum-structured latent states provide a strong inductive bias for learning local cellular-automaton dynamics.

\mypara{Finding 2: The improvement is robust to capacity-controlled baselines.}
ComplexQSWM outperforms the parameter-matched classical baseline, the doubled-latent real-valued control, and the normalized real-valued control in one-step prediction.
Because these baselines control for model size, real-valued latent capacity, and normalization, the result points to the structure of the complex-valued latent representation as an important factor.
ComplexQSWM also achieves the best short-horizon rollout accuracy at horizon $3$, further supporting its advantage for local predictive dynamics.

\begin{figure}[t]
\centering
\begin{minipage}[t]{0.48\linewidth}
\centering
\includegraphics[width=\linewidth]{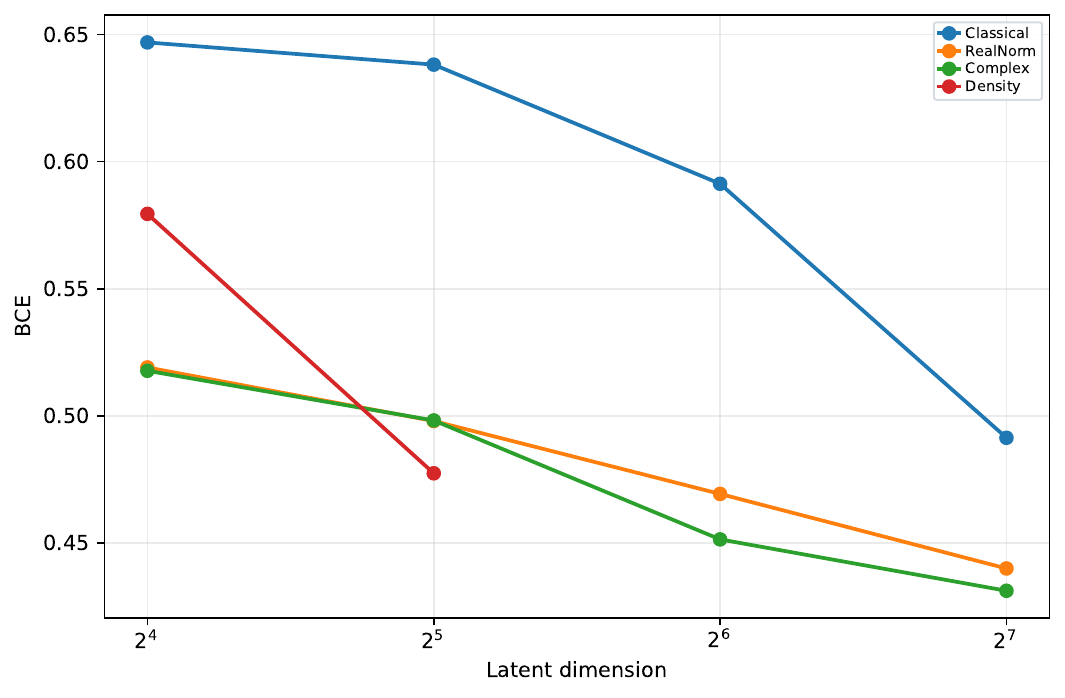}
\par\smallskip\small (a) Scaling over latent dimension.
\end{minipage}
\hfill
\begin{minipage}[t]{0.48\linewidth}
\centering
\includegraphics[width=\linewidth]{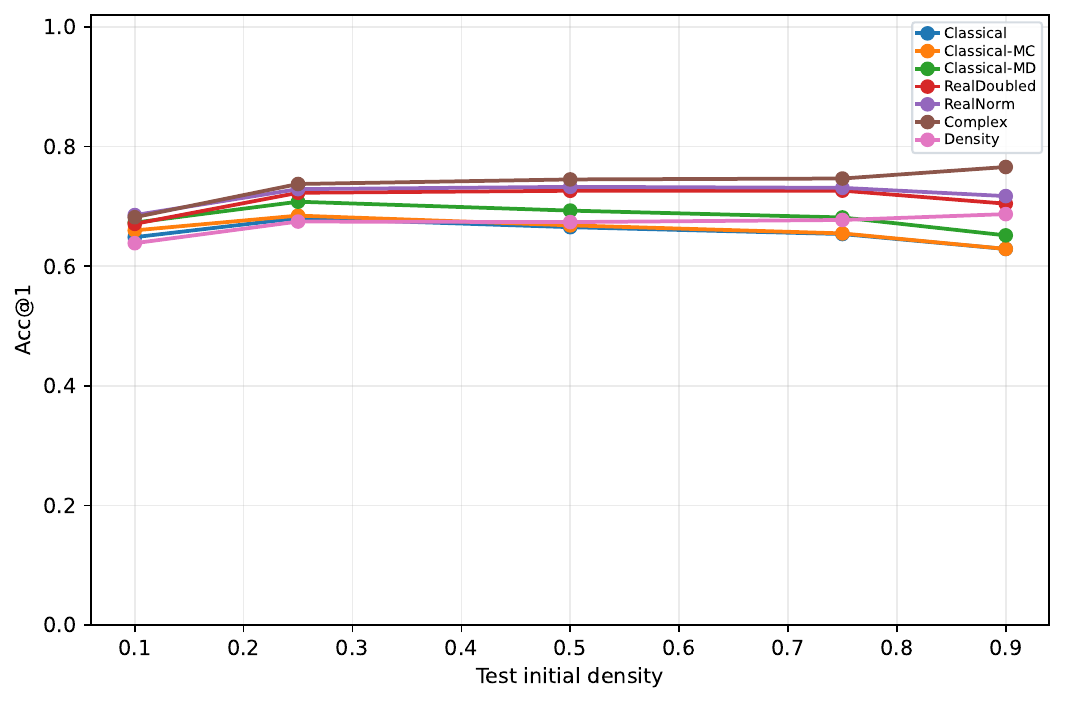}
\par\smallskip\small (b) OOD initial-density shift.
\end{minipage}
\caption{
Scaling and out-of-distribution density evaluation.
(a) One-step BCE under increasing latent dimension.
(b) One-step accuracy under initial-density shifts from the training density $p=0.5$.
}
\label{fig:scaling_ood}
\end{figure}

\mypara{Finding 3: ComplexQSWM remains robust across settings.}
Figure~\ref{fig:scaling_ood} evaluates whether the benefit of ComplexQSWM persists beyond a single fixed setting.
The scaling study shows that ComplexQSWM maintains favorable predictive trends as latent dimension increases.
The OOD density experiment further shows competitive performance when the initial-state distribution shifts away from the training density $p=0.5$.
Together, these results suggest that quantum-structured latent representations provide a useful inductive bias across model capacity and data-distribution changes.

\begin{figure}[t]
\centering
\begin{minipage}[t]{0.48\linewidth}
\centering
\includegraphics[width=\linewidth]{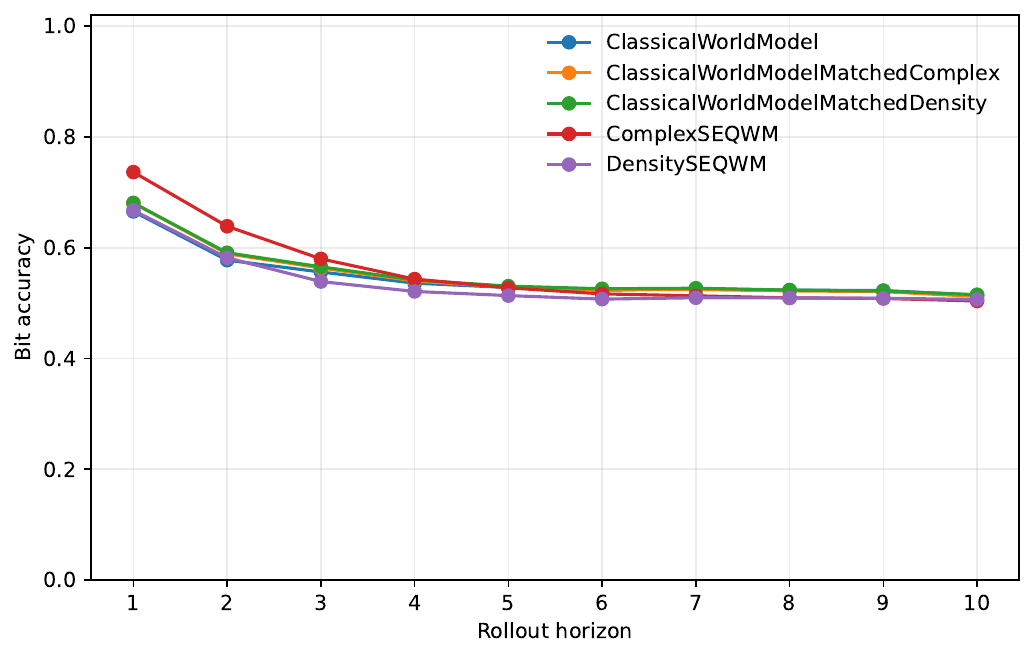}
\end{minipage}
\hfill
\begin{minipage}[t]{0.48\linewidth}
\centering
\includegraphics[width=\linewidth]{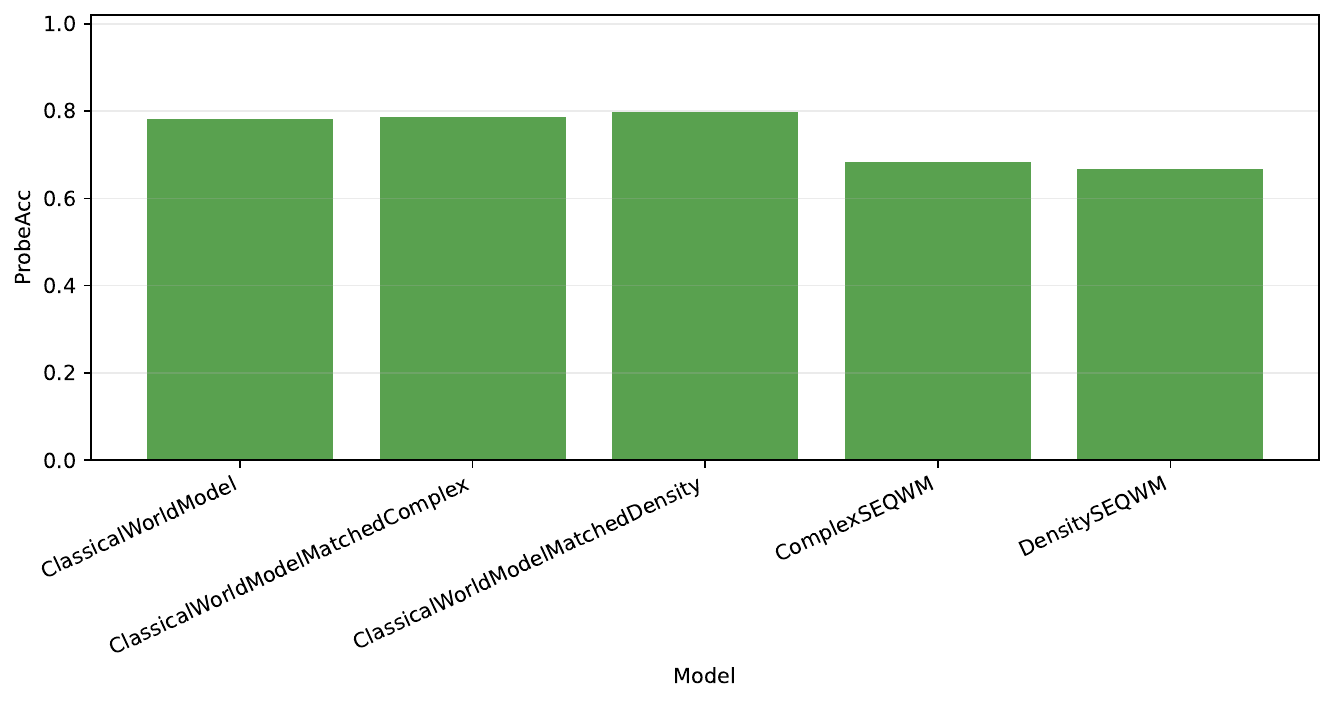}
\end{minipage}
\caption{
\small Recursive rollout accuracy (left) and Latent probing accuracy (right).
}
\label{fig:rollout_probe}
\end{figure}

\mypara{Finding 4: Rollout and probing characterize learned dynamics.}
Figure~\ref{fig:rollout_probe} characterizes the learned latent dynamics beyond one-step prediction.
Rollout measures whether the learned transition remains useful when its own predictions are recursively reused, while probing measures how directly the latent state exposes the underlying cellular-automaton configuration.
The rollout results show that ComplexQSWM has its clearest advantage at local and short horizons.
The probing results show that the strongest predictive model is not necessarily the most linearly decodable.
This suggests that ComplexQSWM improves prediction through structured latent dynamics rather than by simply forming a latent state that is directly readable by simple probes.

With these findings, we answer the central question as:
\begin{graybox}
\textit{\textbf{Yes.} QSWMs provide a useful framework for studying quantum-structured latent dynamics in predictive world modeling. 
ComplexQSWM shows the promise of quantum-structured representations for future studies of emergent world-modeling capabilities.}
\end{graybox}

\section{Discussion and Conclusion}
\label{sec:discussion_conclusion}

This paper introduced \emph{Quantum-Structured World Models} (QSWMs), a framework for predictive world modeling with quantum-structured latent states, latent transition operators, and measurement-inspired decoding.
We formalized QSWMs through history encoding, latent evolution, and predictive readout; established three foundational properties: classical inclusion, predictive sufficiency, and structured compactness; and instantiated ComplexQSWM and DensityQSWM on cellular-automaton dynamics.
The results show that ComplexQSWM achieves the strongest local predictive performance across controlled comparisons, suggesting that complex-valued latent structure can provide useful inductive bias beyond parameter count, doubled real latents, or normalization alone.

Overall, QSWMs provide a principled and testable framework for studying quantum-structured latent representations in world models.
Scaling and OOD-density experiments show that the benefit of ComplexQSWM extends beyond a single fixed setting, while rollout and probing analyses reveal complementary properties of the learned latent dynamics.
These findings support QSWMs as a promising representational foundation for predictive world modeling and for future studies of emergent world-modeling capabilities.

\newpage
\bibliographystyle{splncs04}
\bibliography{reference}
\newpage
\appendix

\section{Detailed Proofs}
\label{app:proofs}

\subsection{Proof of Theorem 1: Classical Inclusion}

\textbf{Theorem 1.}
Every finite-dimensional classical probabilistic latent world model can be represented as a restricted QSWM whose latent states are diagonal density operators and whose transition and measurement-inspired decoding maps reduce to classical stochastic maps.

\textbf{Proof.}
Consider a finite classical latent state space
\[
\mathcal{Z}={1,2,\ldots,n}.
\]
A classical probabilistic world model maintains a belief vector
\[
p_t=(p_t(1),p_t(2),\ldots,p_t(n)),
\]
where $p_t(i)\geq 0$ and $\sum_i p_t(i)=1$.
Let $\mathcal{H}$ be an $n$-dimensional Hilbert space with computational basis $\{|i\rangle\}_{i=1}^n$.
We embed the classical belief vector into a diagonal density operator:
\[
\rho_t=\sum_{i=1}^{n}p_t(i)|i\rangle\langle i|.
\]
This operator is positive semidefinite and has unit trace, so $\rho_t\in\mathcal{D}(\mathcal{H})$.

Let $P_a(j\mid i)$ be a classical action-conditioned transition matrix.
The updated belief is
\[
p_{t+1}(j)=\sum_{i=1}^{n}P_a(j\mid i)p_t(i).
\]
Define a QSWM transition map $\mathcal{T}_a$ that acts on diagonal density operators as
\[
\mathcal{T}_a(\rho_t)
=
\sum_{j=1}^{n}
\left(
\sum_{i=1}^{n}P_a(j\mid i)p_t(i)
\right)
|j\rangle\langle j|.
\]
This map preserves positivity, trace, and diagonality, and exactly reproduces the classical belief update.

Similarly, let $O(o\mid j)$ be a classical observation model.
Define a measurement-inspired decoding map by
\[
p(o\mid \rho_{t+1})
=
\sum_{j=1}^{n}
O(o\mid j)\langle j|\rho_{t+1}|j\rangle.
\]
Since $\langle j|\rho_{t+1}|j\rangle=p_{t+1}(j)$, this map produces the same predictive distribution as the classical observation model:
\[
p(o\mid \rho_{t+1})
=
\sum_{j=1}^{n}O(o\mid j)p_{t+1}(j).
\]
Thus, every finite-dimensional classical probabilistic latent world model is recovered by restricting QSWMs to diagonal latent operators, stochastic transition maps, and classical measurement-inspired decoding maps.
\hfill $\square$

\subsection{Proof of Theorem 2: Predictive Sufficiency}

\textbf{Theorem 2.}
Assume that a QSWM achieves optimal prediction of a future target $y_t$ from history $h_t$ under a strictly proper predictive loss.
Then any minimal optimal latent state $\rho_t=E_\theta(h_t)$ is a predictive sufficient statistic:
\[
p(y_t\mid h_t)=p(y_t\mid \rho_t).
\]

\textbf{Proof.}
A strictly proper predictive loss is minimized when the predicted distribution equals the true conditional distribution of the target.
Thus, an optimal predictor given the full history recovers
\[
p(y_t\mid h_t).
\]
In a QSWM, prediction is made only through the latent state
\[
\rho_t=E_\theta(h_t).
\]
Therefore, the predicted distribution has the form
\[
q_\theta(y_t\mid \rho_t).
\]
Optimality implies that, for every history $h_t$,
\[
q_\theta(y_t\mid E_\theta(h_t))=p(y_t\mid h_t).
\]

Now suppose two histories $h_t$ and $h'*t$ are mapped to the same latent state:
\[
E*\theta(h_t)=E_\theta(h'*t)=\rho.
\]
Since the prediction depends only on $\rho$, the QSWM assigns the same predictive distribution to both histories:
\[
q*\theta(y_t\mid E_\theta(h_t))
=
q_\theta(y_t\mid E_\theta(h'_t)).
\]
By optimality,
\[
p(y_t\mid h_t)=p(y_t\mid h'_t).
\]
Therefore, the encoder can merge two histories only when they induce the same future predictive distribution.

A minimal optimal latent state therefore preserves exactly the predictive equivalence classes of histories needed for predicting $y_t$.
Equivalently, conditioning on $\rho_t$ makes the full history unnecessary for predicting the future target:
\[
p(y_t\mid h_t)=p(y_t\mid \rho_t).
\]
Thus, $\rho_t$ is a predictive sufficient statistic.
\hfill $\square$

\subsection{Proof of Theorem 3: Structured Compactness}

\textbf{Theorem 3.}
Consider a family of environments whose predictive state distributions over $n$ latent factors admit tensor-network representations with bond dimension at most $\chi$.
A QSWM using tensor-structured latent states can represent these predictive states with $O(n\chi^2)$ parameters, while an unstructured classical state table requires $O(2^n)$ parameters.

\textbf{Proof.}
Consider $n$ binary latent factors
\[
x=(x_1,\ldots,x_n)\in{0,1}^n.
\]
A general unstructured classical probability table assigns one probability to every configuration $x$.
Since there are $2^n$ configurations, this representation requires
\[
O(2^n)
\]
parameters, up to the normalization constraint.

Now suppose the predictive state distribution admits a tensor-network representation with bond dimension at most $\chi$.
For example, in a one-dimensional matrix-product representation, the predictive tensor can be written as
\[
P(x_1,\ldots,x_n)
=
\mathrm{Tr}
\left(
A^{[1]}_{x_1}
A^{[2]}_{x_2}
\cdots
A^{[n]}_{x_n}
\right),
\]
where each local tensor $A^{[i]}_{x_i}$ has internal bond dimension at most $\chi$.
For binary variables, each site has a constant number of local values, and each local tensor contributes $O(\chi^2)$ parameters.
Thus, the total number of parameters scales as
\[
O(n\chi^2).
\]
If $\chi$ is constant or polynomial in $n$, this representation is polynomial in $n$.

A QSWM can use such tensor-structured latent states as its quantum-structured representation.
Therefore, for predictive distributions with bounded-bond-dimension tensor structure, QSWMs can represent the predictive state with $O(n\chi^2)$ parameters, while an unstructured classical state table requires $O(2^n)$ parameters.
\hfill $\square$

\section{Implementation Details}
\label{app:implementation}

This appendix provides implementation details for the experimental validation.

\subsection{Dataset Generation}

We use one-dimensional elementary cellular automata with binary states of length $L=32$.
Each rule updates a cell from its left, center, and right neighbors under periodic boundary conditions.
For a neighborhood $(x_{j-1},x_j,x_{j+1})$, the corresponding three-bit pattern indexes the rule table.

For each rule, random initial states are sampled from a Bernoulli distribution.
Each trajectory is generated for $T=20$ steps.
A training example consists of a history window of length $H=4$ and the next target state.
Unless otherwise stated, each rule uses $5000$ training trajectories, $1000$ validation trajectories, and $1000$ test trajectories.
We evaluate Rule 30, Rule 90, and Rule 110, which represent chaotic-looking, linear/fractal, and complex cellular-automaton dynamics, respectively.

\subsection{Training Protocol}

All models are trained for $20$ epochs using Adam with learning rate $10^{-3}$ and batch size $128$.
The loss is binary cross entropy over the next-state bits.
The best validation checkpoint is used for test evaluation.
All experiments use fixed random seeds for Python, NumPy, and PyTorch.

\subsection{Recursive Rollout Evaluation}

For rollout evaluation, the model is initialized with the first $H$ ground-truth states.
It predicts the next state, applies a sigmoid function, thresholds each bit at $0.5$, and appends the predicted state to the history.
The model then recursively predicts future states using its own previous predictions.
We report bit accuracy at horizons $1$, $3$, $5$, and $10$.
This protocol evaluates the stability of learned transition dynamics under recursive prediction.

\subsection{Latent Probing}

To evaluate whether learned latent states contain recoverable world-state information, each trained world model exposes an encoder function,
\[
\mathrm{encode_latent}(h_t).
\]
The world model is frozen, and a probe is trained to predict the current cellular-automaton state, defined as the last state in the input history.
The probe is trained for $10$ epochs using Adam with learning rate $10^{-3}$ and binary cross entropy.

For classical baselines, the probe input is the classical latent vector.
For ComplexQSWM, the probe input is the concatenation of the normalized real and imaginary latent components.
For DensityQSWM, the probe input is the flattened pre-transition density-matrix-like latent state.
Probe performance is reported as bit accuracy.

\section{Per-Rule Results}
\label{app:per_rule}

Table~\ref{tab:per_rule_results} reports full per-rule results for Rule 30, Rule 90, and Rule 110.
The aggregate results in the main paper are computed by averaging across these three rules.
ComplexQSWM shows the strongest one-step prediction on Rule 30 and Rule 110, while all models perform similarly on Rule 90 under the current setup.
This indicates that the benefit of complex-valued quantum-structured latents is dynamics-dependent rather than uniform across all cellular-automaton rules.

\begin{table}[t]
\centering
\caption{Per-rule results for elementary cellular-automaton world-modeling tasks. Lower BCE is better; higher accuracy is better.}
\label{tab:per_rule_results}
\scriptsize
\setlength{\tabcolsep}{3.2pt}
\begin{tabular}{llrrrrrr}
\toprule
Rule & Model & BCE $\downarrow$ & Acc@1 $\uparrow$ & Acc@3 $\uparrow$ & Acc@5 $\uparrow$ & Acc@10 $\uparrow$ & Probe $\uparrow$ \\
\midrule
30 & ClassicalWM & 0.5267 & 0.7388 & 0.6021 & 0.5295 & 0.5137 & 0.9423 \\
30 & Matched-Complex & 0.5075 & 0.7457 & 0.6061 & 0.5285 & 0.5108 & 0.9479 \\
30 & Matched-Density & 0.4809 & 0.7561 & \textbf{0.6143} & \textbf{0.5337} & \textbf{0.5148} & \textbf{0.9678} \\
30 & ComplexQSWM & \textbf{0.3911} & \textbf{0.8238} & 0.6104 & 0.5263 & 0.4999 & 0.8318 \\
30 & DensityQSWM & 0.5562 & 0.7022 & 0.5319 & 0.4998 & 0.5056 & 0.7279 \\
\midrule
90 & ClassicalWM & 0.6854 & 0.5627 & 0.4989 & \textbf{0.5086} & 0.4913 & 0.5883 \\
90 & Matched-Complex & 0.6859 & \textbf{0.5677} & 0.4998 & 0.5084 & 0.4913 & 0.5810 \\
90 & Matched-Density & 0.6856 & 0.5669 & \textbf{0.5000} & \textbf{0.5086} & 0.4913 & 0.5864 \\
90 & ComplexQSWM & \textbf{0.6817} & 0.5652 & 0.4990 & 0.5084 & 0.4913 & 0.5191 \\
90 & DensityQSWM & 0.6927 & 0.5628 & 0.4996 & 0.5084 & 0.4913 & \textbf{0.6144} \\
\midrule
110 & ClassicalWM & 0.5685 & 0.6947 & 0.5674 & 0.5462 & 0.5381 & 0.8129 \\
110 & Matched-Complex & 0.5280 & 0.7286 & 0.5841 & 0.5485 & 0.5336 & 0.8310 \\
110 & Matched-Density & 0.5377 & 0.7189 & 0.5820 & \textbf{0.5496} & \textbf{0.5392} & \textbf{0.8423} \\
110 & ComplexQSWM & \textbf{0.3942} & \textbf{0.8203} & \textbf{0.6301} & 0.5477 & 0.5200 & 0.6963 \\
110 & DensityQSWM & 0.5143 & 0.7369 & 0.5849 & 0.5323 & 0.5227 & 0.6604 \\
\bottomrule
\end{tabular}
\end{table}

\section{Training Curves}
\label{app:curves}

\begin{figure*}[t]
\centering
\includegraphics[width=0.9\textwidth]{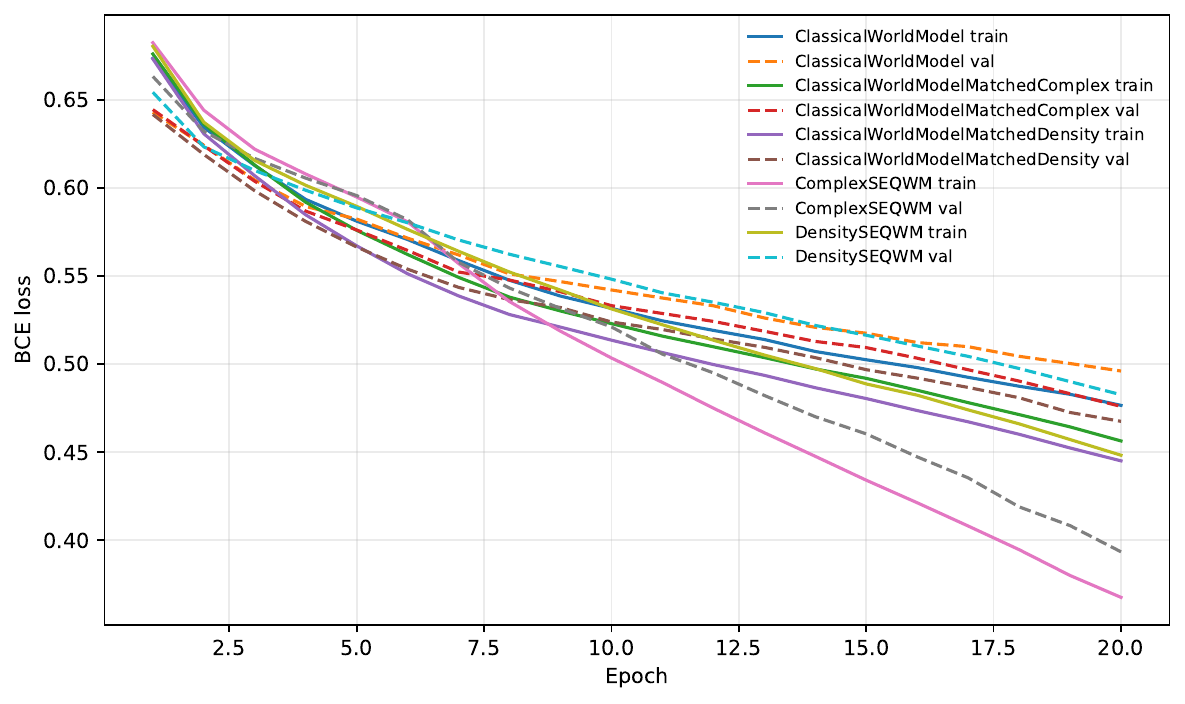}
\caption{Training and validation BCE curves for all evaluated models. ComplexQSWM achieves the lowest validation loss by the end of training.}
\label{fig:training_curve}
\end{figure*}

Figure~\ref{fig:training_curve} shows the training and validation BCE curves for all evaluated models under the same training protocol.
All models exhibit decreasing training loss, indicating stable optimization.
ComplexQSWM achieves the lowest validation BCE by the end of training, providing supporting evidence for its stronger local predictive performance in the main results.

\section{Latent Probing Results}
\label{app:probe}

We further evaluate whether learned latent states contain recoverable information about the current cellular-automaton state.
For each trained model, we freeze the world model and train a linear probe to predict the current state, defined as the last state in the input history.

Table~\ref{tab:probe} reports aggregate probing results.
The classical matched baselines achieve higher linear probing accuracy than the QSWM variants.
This indicates that the predictive gain of ComplexQSWM is not simply due to a more linearly decodable latent state.
Instead, ComplexQSWM may encode information in a form that is useful for local predictive dynamics but less directly accessible to a linear probe.

\begin{table}[t]
\centering
\small
\caption{Aggregate latent probing results. Lower ProbeBCE is better; higher ProbeAcc is better.}
\label{tab:probe}
\begin{tabular}{lcc}
\toprule
Model & ProbeBCE $\downarrow$ & ProbeAcc $\uparrow$ \\
\midrule
ClassicalWM & 0.5024 & 0.7812 \\
Matched-Complex & 0.4910 & 0.7866 \\
Matched-Density & 0.4686 & 0.7988 \\
ComplexQSWM & 0.6392 & 0.6824 \\
DensityQSWM & 0.6235 & 0.6676 \\
\bottomrule
\end{tabular}
\end{table}

\begin{figure}[t]
\centering
\includegraphics[width=0.85\linewidth]{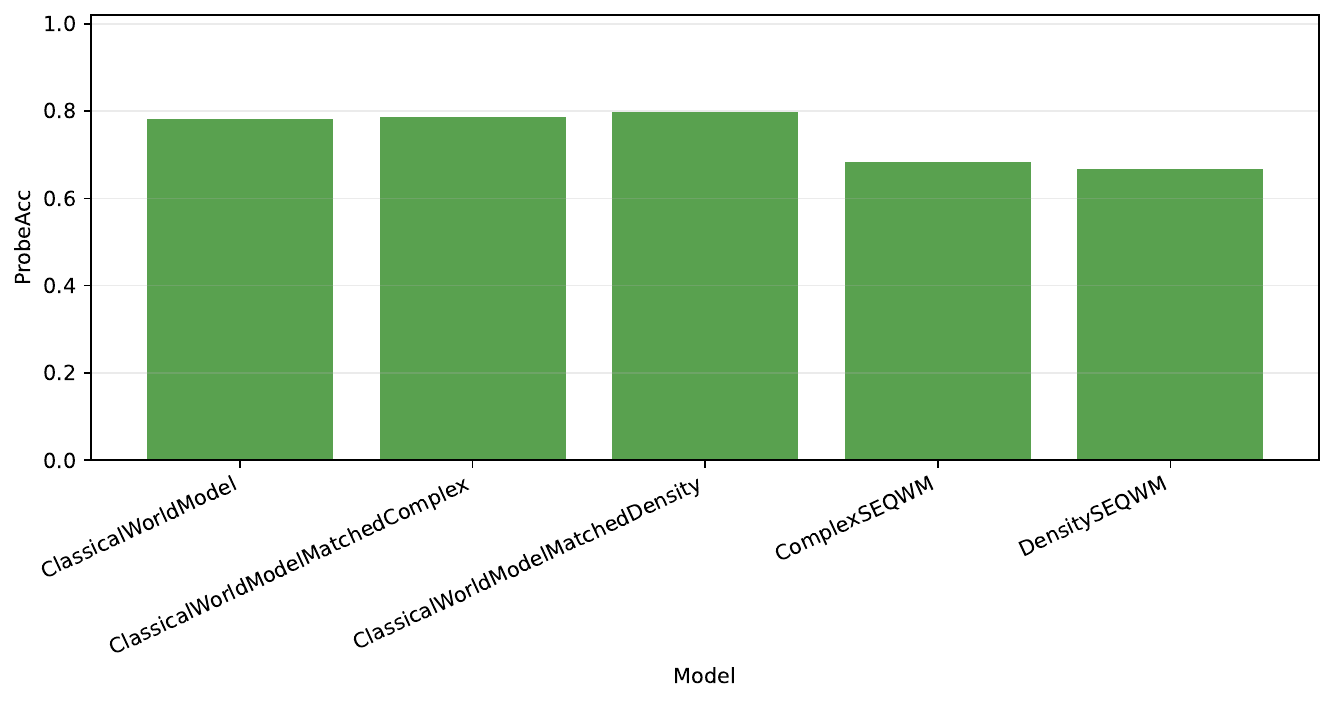}
\caption{Linear probing accuracy of learned latent states.}
\label{fig:latent_probe_accu}
\end{figure}

\section{Additional QSWM Architectures}
\label{app:architectures}

The main paper focuses on quantum-inspired QSWMs implemented with classical differentiable architectures.
The same framework also suggests two future architectural directions: hybrid quantum-classical QSWMs and quantum-native QSWMs.

\mypara{Hybrid quantum-classical QSWMs.}
A hybrid QSWM keeps high-dimensional history encoding and output decoding in classical neural networks, while implementing selected latent modules with parameterized quantum circuits.
For example, a classical encoder may compress the history into features $x_t=C_\theta(h_t)$, which are then used to prepare a quantum latent state $|\psi_t\rangle=\mathrm{Prep}(x_t)$.
An action-conditioned quantum circuit can evolve the latent state,
\[
|\psi_{t+1}\rangle = U_\theta(a_t)|\psi_t\rangle,
\]
and measurement statistics can be passed to a classical decoder for prediction.
This design provides a natural intermediate step between classical quantum-inspired QSWMs and fully quantum-native world models.

\mypara{Quantum-native QSWMs.}
A quantum-native QSWM represents and evolves the latent world state directly as a quantum state or density operator,
\[
\rho_t \in \mathcal{D}(\mathcal{H}),
\qquad
\rho_{t+1}=\mathcal{E}_{a_t}(\rho_t),
\]
where $\mathcal{E}_{a_t}$ is an action-conditioned quantum channel.
Predictions are obtained through measurements,
\[
p(y=m\mid \rho_t)=\mathrm{Tr}(M_m\rho_t).
\]
Such models are especially natural for quantum physical environments, including spin systems, quantum circuits, and quantum control tasks.

\begin{table}[t]
\centering
\caption{Representative QSWM architectural classes.}
\label{tab:architectures}
\small
\setlength{\tabcolsep}{4pt}
\begin{tabular}{lll}
\toprule
Architecture & Latent state & Role \\
\midrule
Quantum-inspired & Complex or density-like states & Classical validation \\
Hybrid quantum-classical & Quantum circuit modules & Near-term quantum extension \\
Quantum-native & Physical quantum states/channels & Quantum physical modeling \\
\bottomrule
\end{tabular}
\end{table}

\section{Future Evaluation Directions}
\label{app:evaluation}

The main experiments use elementary cellular automata because they provide controlled latent dynamics and exact transition rules.
Future QSWM studies can extend evaluation to broader environments.

\mypara{Symbolic and programmatic environments.}
Grid worlds, finite-state machines, board games, and programmatic environments provide hidden states and exact transition rules.
They can test whether QSWM latent states capture world variables that are useful for prediction, planning, and counterfactual reasoning.

\mypara{Physical simulation.}
Continuous dynamical systems, particle interactions, low-dimensional robotics tasks, and locally interacting physical systems can test whether QSWMs model structured uncertainty and multi-step dynamics.
Evaluation can include one-step error, rollout error, conservation consistency, and robustness under distribution shift.

\mypara{Quantum physical systems.}
Spin chains, quantum circuits, and quantum control environments are natural benchmarks for quantum-structured world models.
In such settings, evaluation may include state fidelity, trace distance, expectation-value error, and measurement-distribution divergence.

\mypara{Planning and control.}
QSWMs can also be evaluated as planning models by recursively simulating candidate future trajectories,
\[
\rho_{t+j}
=
\mathcal{T}_{\theta,a_{t+j-1}}(\rho_{t+j-1}),
\qquad j=1,\ldots,k,
\]
and scoring them with a reward or value decoder.
Planning-oriented evaluation can include cumulative reward, success rate, rollout depth, and robustness to compounding prediction errors.

\end{document}